\RequirePackage{amsmath,amssymb,amsbsy,amsfonts}
\RequirePackage{algorithm,algpseudocode}
\documentclass[runningheads]{llncs}
 \usepackage[T1]{fontenc}
 \usepackage{lmodern}
 \usepackage{todonotes}
 \usepackage{verbatimbox}
 \usepackage{graphicx}
\usepackage{tikz, pgfplots}
\usetikzlibrary{arrows}
\usepackage{subfig}
\usepackage{bibnames}
\usepackage{tablefootnote}

\begin{document}
 \title{Neural Nets with a Newton Conjugate Gradient Method on Multiple GPUs  }
\titlerunning{Newton-CG for large ResNets} 
\author{Severin Reiz, Tobias Neckel, Hans-Joachim Bungartz}
\institute{Technical University of Munich (TUM), School of  Computation, Information and Technology}
\maketitle

\abstract{Training deep neural networks consumes increasing computational resource shares in many compute centers. Often, a brute force approach to obtain hyperparameter values is employed. Our goal is (1) to enhance this by  enabling second-order optimization methods with fewer hyperparameters for large-scale neural networks and (2) to perform a survey of the performance optimizers for specific tasks to suggest users the best one for their problem. We introduce a novel second-order optimization method that requires the effect of the Hessian on a vector only and avoids the huge cost of explicitly setting up the Hessian for large-scale networks.

We compare the proposed second-order method with two state-of-the-art optimizers on five representative neural network problems, including regression and very deep networks from computer vision or variational autoencoders. For the largest setup, we efficiently parallelized the optimizers with Horovod and applied it to a 8 GPU NVIDIA P100 (DGX-1) machine. }  
 \keywords{Numerical methods;
 Machine learning;
 Deep learning;
 Second-order optimization;
 Data parallelism}



\section{Introduction}


Machine Learning (ML) is widely used in todays software world: regression or classification problems are solved to obtain efficient models of input-output relationships learning from measured or simulated data. In the context of scientific computing, the goal of ML frequently is to create surrogate models of similar accuracy than existing models but with evaluation runtimes of much cheaper computational costs.
Applying an ML technique typically results in an online vs.~an offline phase. While the offline phase comprises all computational steps to create the ML model from given data (the so-called training data), the online phase is associated to obtaining desired answers for new data points (typically called validation points).
Different types of ML techniques exist: Neural networks in various forms, Gaussian processes which incorporate uncertainty, etc. \cite{Goodfellow-et-al-2016}. Particularly prominent are deep learning models which are commonly used for a huge variety of applications, notably in safety-critical fields such as autonomous driving, natural language processing and medical image processing and ML for scientific computing.

For almost all methods, a numerical optimization is necessary to tune parameters or hyperparameters of the corresponding method in the offline phase such that good/accurate results can be obtained in the online phase. 
  Even though numerical optimization is a comparably mature field that offers many solution approaches, the optimization problem associated with real-world large-scale ML scenarios is non-trivial and computationally very demanding: The dimensionality of the underlying spaces is high, the amount of parameters to be obtimised is large to enormous, and cost function---frequently called loss function---is typically mathematically complicated being non-convex and possessing many local optima and saddle points in general. 
  Additionally, the performance of a method typically depends not only on the ML approach but also on the scenario of application.

Of the zoo of different optimization techniques, certain first-order methods such as the stochastic gradient descent (SGD) have been very popular and represent the de-facto fallback in many cases.
  Higher order methods provide generally nice convergence features since they include curvature information of the loss function to be optimized to avoid unnecessary  optimization steps. These methods, however, come at the price of evaluating the Hessian of the problem, which typically is way too costly for real-world large-scale ML scenarios, both w.r.t.~setting up and storing the matrix and w.r.t.~evaluating the matrix-vector product with standard implementations (e.g., the ResNet50 scenario discussed below has about 16 million degrees of freedom in form of corresponding weights). 
  In this paper, we will analyze a second-order Newton-based optimization method w.r.t.~accuracy and computational performance in the context of large-scale neural networks of different type. To cope with challenging costs in such scenarios, we implemented a special variant of a regularized Newton method using the Pearlmutter scheme together with a matrix-free conjugate gradient method which results in comparably low cost to evaluate the effect of the Hessian on a given vector of about twice the cost of a backpropagation itself.
  
  All implementations are publicly available and easy to integrate since they rely on TensorFlow Keras code.
  We are comparing our proposed solution with existing TensorFlow implementations of the prominent SGD and Adam method for five representative ML scenarios of different categories to get insight into the question when which optimizer will be advantageous.
  In particular, we exploit in our implementation parallelisation in the optimization process on two different levels: a parallel execution of runs the as well as data parallelism by treating several chunks of data (the so-called batches or mini batches) in parallel, too. The latter results in a quasi-Newton method where the effect of the Hessian is kept constant for a couple of data points before the next update is computed. Our approach is therefore a combination of usability, accuracy and efficiency.

    

The remainder of this paper is organized as follows. Section \ref{sec:related} lists work in the community that is related to our approach.
In Sec.~\ref{sec:methods}, basic aspects of deep neural networks are briefly stated to fix the nomenclature for the algorithmic building blocks we combine for our method. The detailed neural network structures and architectures for the five scenarios to be discussed are discussed in Sec.~\ref{sec:arch}.
We briefly describe aspects of the implementation in Sec.~\ref{sec:implementation} and show results for the five neural network scenarios in Sec.~\ref{sec:results}. 
Section~\ref{sec:conclusion} finally concludes the discussion.

\section{Related work}\label{sec:related}

Hessian multiplication for neural networks without forming the matrix was introduced very early~\cite{pearlmutter1994fast}; while there are multiple optimization techniques around~\cite{nocedal1999numerical}, it gained importance again with \textit{Deep Learning via Hessian-free optimization} \cite{martens2010deep}. Later, the Kronecker-factored approximate curvature (KFAC) of the Fisher matrix(similar to Gauss-Newton Hessian) was introduced~\cite{martens2016second}; for \textit{high performance computing}, chainerkfac was introduced~\cite{osawa2019large}. AdaHessian uses the Hutchinson method for adapting learning rate~\cite{yao2020adahessian}, other work involves inexact newton methods for neural networks~\cite{o2019inexact} or a comparison of optimizers~\cite{schmidt2021descending}. With GOFMM, we performed initial studies on Hessian matrices~\cite{chenhan2018distributed}, where later we looked at the fast approximation for a multilayer perceptron~\cite{chen2021fast}.

%
%
%
%
%
%

\section{Methods}\label{sec:methods}


In this section, we first briefly describe the basics of deep neural networks\footnote{For a brief introduction on deep NN, cf.~\cite{Lecun2015}} and the pecularities of the variants we are going to use in the five different scenarios in Sec.~\ref{sec:results}. Afterwards, we highlight the basic algorithmic ingredients of the reference implementations (SGD and Adam)~\cite{Goodfellow-et-al-2016}. Finally, we explain the building blocks of our approach: The Pearlmutter trick and the Newton-CG step.


\subsection{Scientific Computing for Deep Learning} \label{sec:DL}

Consider a feed-forward deep neural network defined as the parameterized function $f (X, \textbf{W} )$. The function $f$ is composed by vector-valued functions $f^{(i)}, \ i=1,\ldots,D$, which represent each one layer in the network of depth $D$, in the following way: $f (x) = f^{(D)}( \dots f^{(2)}( f^{(1)} (x)) )$


The function corresponding to a network layer \((d)\) and the output of the j-th neuron are computed via

$$f^{(d)} = \begin{bmatrix} z_1^{(d)} \\ z_2^{(d)} \\ \vdots\\ z^{(d)}_{M^{(d)}}\end{bmatrix}
\mbox{and}\ z_j^{(d)}= \phi( \sum_{i=1}^{M^{(d-1)}}(
w_{ji}^{(d)} f_i^ {(d-1)})
+ w_j^0 )$$
with activation function $\phi$ and weights $w$. All weights $w$ are comprised in a large vector $\textbf{W}\in\mathbb{R}^n$ which represents a parameter for $f$.
The optimization problem consists now of finding weights $\textbf{W}$ a given loss function $l$ will be minimized for given training samples $X,Y$:
\begin{align}
\min_{\textbf{W}} l(X, Y, \textbf{W}) \label{eq:minProblem}
\end{align}
A prominent example of a loss function is the categorical cross-entropy 
$$
l_{entr} (X, Y, \textbf{W}) := -\sum\limits_{l=1}^{\texttt{N}}y_i \log(f^{(D)}(X,\textbf{W}) ) \ .
$$
Note that only the last layer function $f^{(D)}$ of the network directly shows up in the loss, but all layers are indirectly relevant due to the optimization for all weights in all layers.

Optimizers look at stochastic \textit{mini batches} of data, i.e.~disjoint collections of data points. The union of all mini batches will represent the whole training data set. The reason for considering data in chunks of mini batches and not in total is that the backpropagation in larger neural networks will face severe issues w.r.t. memory. 
Hence, the mini batch loss function is now defined by
$$
L (x, y, \textbf{W}) := -\sum\limits_{i=1}^{\textnormal{batch-size}} y_i \log(f^{(D)}(X,\textbf{W}) ) \ ,
$$
where the mini batch is varied in each optimization step in a round-robin manner.

\subsection{State-of-the-art Optimization Approaches} 
In order to solve the optimization problem \eqref{eq:minProblem}, different first-order methods exist (for a survey, see \cite{Goodfellow-et-al-2016}, e.g.). The pure \textit{gradient descent} without momentum is computing weights \(\textbf{W}_k\) in iteration $k$ via \(\textbf{W}_k = \textbf{W}_{k-1} - a_{k-1} \nabla l(\textbf{W}_{k-1} ) \) where \( \nabla L(\textbf{W}_{k-1} ) \) denotes the gradient of the total loss function $l$ w.r.t.~the weights $\textbf{W}$.

The \textit{stochastic gradient descent} (SGD) includes stochasticity by changing the loss function to the input of a specific mini batch of data, i.e.~using $L$ instead of $l$. Each mini batch of data provides a noisy estimator of the average gradient over all data points, hence the term stochastic.
Technically, this is realised by switching the mini batches in a round-robin manner to reach over the full dataset (one full sweep is called an epoch; frequently, more than one epoch of iterations is necessary to achieve quality in the optimization).

The family of \textit{Adam} methods updates weight values by moving averages of the gradient \(s_k\) with estimates of the \(1^{st}\) moment (the mean) and the \(2^{nd}\) raw moment (the uncentered variance)

The approach called \textit{AdaGrad} is directly using these estimators: 
\begin{align}
    W_{k+1} = W_{k} - \alpha_k  \frac{s_k}{\delta+\sqrt{r_k}} \label{eq:adagrad}
\end{align} 
The \textit{Adam} method corrects for the biases in the estimators by using the estimators \(\hat{s_k}=\frac{s_k}{1-\beta_1^k}\) and \(\hat{r_k}=\frac{r_k}{1-\beta_2^k}\) instead of \(s_k\) and \(r_k\). Good default settings for the tested machine learning problems described in this paper are \(a_0 = 0.001\), \(\beta_1 = 0.9\), \(\beta_2 = 0.999\), and \(\delta = 10^{-8}\) 

\subsection{Proposed 2nd-order Optimizer}

The second-order optimizer implemented and used for the results of this work consists of a Newton scheme with a matrix-free conjugate gradient (CG) solver for the linear systems of equations arising in each Newton step. The effect of the Hessian on a given vector (i.e.~a matrix-vector multiplication result) is realised via the so-called Pearlmutter approach and avoids setting up the Hessian explicitly.

\subsubsection{Pearlmutter Approach}
The explicit setup of the Hessian is memory-expensive due to the quadratic dendence on the problem size; e.g., a 16M\(\times\)16M matrix requires about 1 TB of memory.
We can obtain ``cheap'' access to the problems curvature information by computing the Hessian-vector product.
This method is called Fast Exact Multiplication by the Hessian $H$ (see \cite{pearlmutter1994fast}, e.g.). Specifically, it computes
the Hessian vector product \(Hs\) for any \(s\) in just \textbf{two} (instead of the number of weights \(n\)) 
backpropagations (i.e.~automatic differentiations for 1st derivative components).

For our formulation of the problem it is defined as:
\[
H_L(\textbf{W})s = \begin{pmatrix}\sum\limits_{i=1}^{n} s_i \frac{\delta^2}{\delta w_1\delta w_i} L(\textbf{W})\\
\sum\limits_{i=1}^{n} s_i \frac{\delta^2}{\delta w_2\delta w_i} L(\textbf{W})\\
\vdots\\
\sum\limits_{i=1}^{n} s_i \frac{\delta^2}{\delta w_n\delta w_i} L(\textbf{W})
 \end{pmatrix}
 =  
 \begin{pmatrix}
 \frac{\delta}{\delta w_1}\sum\limits_{i=1}^{n} s_i \frac{\delta}{\delta w_i} L(\textbf{W})\\
  \frac{\delta}{\delta w_2}\sum\limits_{i=1}^{n} s_i \frac{\delta}{\delta w_i} L(\textbf{W})\\
  \vdots\\
   \frac{\delta}{\delta w_n}\sum\limits_{i=1}^{n} s_i \frac{\delta}{\delta w_i} L(\textbf{W})
\end{pmatrix}
= \nabla_w ( \nabla_w L(\textbf{W}) \cdot s)
\]
The Pearlmutter approach in algorithmic form is shown in Algorithm \ref{alg:pearlm}.
\begin{algorithm}
\caption{\label{alg:pearlm}Pearlmutter}
\begin{algorithmic}[1]
\Require{$X,Y,\textbf{W},s$: \quad Compute $H_L s = \nabla_w ( \nabla_w L(\textbf{W}) \cdot s)$}
\Require{$W_0$: \quad Initial estimate for $\textbf{W}$.}
\State $g_0 \gets \mbox{gradient}( L(\textbf{W}))$ \Comment{Back-Prop}
	\State $\mbox{intermediate} \gets \mbox{matmul}(g_0,s) $ \Comment{Matrix-Multiplication}
	\State $Hs \gets \mbox{gradient}(\mbox{intermediate}) $ \Comment{Back-Prop}
 \State \textbf{return} $Hs$ 
\end{algorithmic}
\end{algorithm}

The resulting formula is both efficient and numerically stable \cite{pearlmutter1994fast}. We denote this building block as \textbf{pearlmutter} in the implementation.



\subsubsection{Newton's method}

Recall the Newton equation
\[H_L(\textbf{W}^k)d^k = -\nabla L(\textbf{W}^k)\]
for the network loss function \(L\) : \(\mathbb{R}^{n} \to \mathbb{R}\), where \(\textbf{W}\) is the vector of network weights and \(\textbf{W}^k\)
the current iterate of Newton's method to solve for the update vector $d^k$. The size of the Hessian is \(\mathbb{R}^{n\times n}\) which becomes infeasible to store with state-of-the-art weight parameter ranges of ResNets (or similar).

Since frequently the Hessian has a high condition number, which implies near-singularity and provokes imprecision, one would apply regularization techniques to counteract a bad condition. A common choice
is Tikhonov regularization. To this end, a multiple of the unit matrix is added to the Hessian of the loss function such that the regularized system is given by
\begin{align}
(H (\textbf{W}^k) + \tau I)d^k = H (\textbf{W}^k )d^k + \tau I d^k = -\nabla L(\textbf{W}^k ) \label{eq:NewtonReg}
\end{align}
Note that for large \(\tau\) the solution will converge to a fraction of the negative gradient
\(\nabla L(\textbf{W}^k )\), similar to a stochastic gradient descent method. The regularized Newton method is summarized in Algorithm \ref{alg:Newton}.
\begin{algorithm}
\caption{\label{alg:Newton}Newton Step}
\begin{algorithmic}[1]
\Require{$\mathbf{W}_0$: \quad Starting point}
\Require{$\tau$: \quad Tikhonov regularization/damping factor}
\State $k \gets 0$
\While{$\mathbf{W}_k$ not converged}
    \State $k \gets k + 1$
    \State $p_k \gets \texttt{CG}(\left(H+\tau I\right), -\nabla L\left(\mathbf{W}_k\right))$ \# Approx $\left(H+\tau I\right)p_k=-\nabla L\left(\mathbf{W}_k\right)$
    \If{$\nabla L\left(\mathbf{W}_k\right)^\top p_k > \tau$} $p_k \gets -\nabla L\left(\mathbf{W}_k\right)$ \# Feasibility check.
   \EndIf
   \State $\alpha_k \gets \alpha$ \# Compute or use a given step size.
   \State $\mathbf{W}_k \gets \mathbf{W}_{k-1} + \alpha_kp_k$
\EndWhile
\end{algorithmic}
\end{algorithm}

\subsubsection{Conjugate gradient step}
Since we have a matrix-vector product available without setting up the full matrix (with \textbf{Pearlmutter}), we employ an iterative solver scheme that requires matrix products only. We therefore employ a few (inaccurate) \texttt{CG}-iterations to solve Newton's regularized equation \eqref{eq:NewtonReg}, thus resulting in an approximated Newton method. The standard \texttt{CG}-algorithm is e.g.~described in~\cite{shewchuk1994introduction}; in particular, no direct matrix-access is required since the algorithm relies only on products with vectors. 


\subsubsection{Complexity}
The method described above requires \(O(bn)\) operations for the evaluation of the gradient, where \(n\) is the number of network weights and \(b\) is the size of the mini batch. In addition,
for the evaluation of the Hessian product and the solution of the Newton-like equation
\(O(2mbn)\) is needed, where \(m\) is the number of iterations conducted by the \texttt{CG} solver
until a sufficient approximation to the solution is reached. Although the second-order optimizer requires more work than ordinary gradient descent, it may still be beneficial since, under the conditions that it promises local q-superlinear convergence,
i.e. \(\exists \gamma \in (0, 1), l \geq 0\), such that
\[||\textbf{W}^{k+1} - \textbf{W}^*|| \leq \gamma || \textbf{W}^k - \textbf{W}^*|| \forall k > l\]
where \(\textbf{W}^\star\) is a local minimizer (see \cite{suk}).


\section{Scenarios and Neural Network architectures} \label{sec:arch}

In this section, we will briefly outline the different neural network structures for the five different ML scenarios used in Sec.~\ref{sec:results}. Those networks share the general structure outlined in Sec.~\ref{sec:DL} but differ in details considerably.
We start with a classical regression case in Sec.~\ref{sec:arch:regr} before briefly explaining the variational autoencoder setting in Sec.~\ref{sec:arch:regr}. The remaining sections deal with classification solutions:  \textit{Bayesian neural networks} are described in Sec.~\ref{sec:arch:bayes}, \textit{convolutional neural networks} in Sec.~\ref{sec:arch:CNN}, and finally basics of \textit{transfer learning} in Sec.~\ref{sec:arch:TL}.

\subsection{Regressional analysis}\label{sec:arch:regr}
Most regression models connect the input \(X\) with some parametric function \(f\) to the output \(Y\), including some error \(\epsilon\), i.e. \(Y=f(X,\beta)+\epsilon\) . The goal is find \(\textbf{W}\) to minimize the loss function which here is the sum of the squared error 
$$
l=\sum_{i=1}^{N}(y_i-f(x_i,\textbf{W}))^2
$$
for all samples $i$ in the training data set.

\subsection{Variational Autoencoder}\label{sec:arch:VAE}
A variational autoencoder (VAE) consists of two coupled but independently parametrized components: The encoder compresses the sampled input \(X\) into the latent space. The decoder receives as input the information sampled from the latent space and produces \({x'}\) as close as possible to \(X\). In a variational autoencoder, encoder and decoder are trained simultaneously such that output \({X'}\) minimizes a reconstruction error to \(X\) by the Kullback-Leibler divergence. For details on VAEs, see \cite{Kingma2019}, e.g.

\subsection{Bayesian Neural Network}\label{sec:arch:bayes}
One of the biggest challenges in all areas of machine learning is deciding on an appropriate
model complexity. Models with too low complexity will not fit the data well, while models possessing high complexity will generalize poorly and provide bad prediction results
on unseen data, a phenomenon widely known as overfitting. Two commonly deployed strategies to counteract this problem are hold-out or cross-validation on one hand, where part of the data is kept from training in order to optimize hyperparameters of the respective model that correspond to model complexity, and controlling the effective complexity
of the model by inducing a penalty term on the loss function on the other hand. The latter approach is known as regularization and can be implemented by applying Bayesian
techniques on neural networks~\cite{bishop1995neural}.

Let \(\theta, \epsilon \sim N (0, 1)\) be random variables, \(w = t(\theta, \epsilon  )\), where \(t\) is a deterministic function.
Moreover, let \(w \sim q(w|\theta)\) be normally distributed. 
Then our optimization task reads
\[l (w, \theta) = \log q(w|\theta) - \log P (D|w) - \log P (w)\]
where the loss function \(l\) is the log-likelihood.~\cite{blundell2015weight}



\subsection{Convolutional Neural Network}\label{sec:arch:CNN}
In general, the convolution is an operation on two functions I, K, defined by 
$$S (t) = (I * K) (t) =\int I (a) K (t - a) da$$
If we use a 2D image I as input
with a 2D kernel K, we obtain a two-dimensional discrete convolution
$$S (i, j) = (I  * K) (i, j)= \sum_x \sum_y I (x, y) K (i - x, j - y)$$

Color images additionally have at least a channel for red, blue and green intensity at each pixel position. Assume that each image is a 3D-tensor and $V_{i,j,k}$ describes the value of channel $i$ at row $j$ and column $k$. Then let our kernel be a 4D-tensor with $K_{l,i,j,k}$ denoting the connection strength (weight) between a unit in input channel $i$ and output channel $l$ at an offset of $k$ rows and $l$ columns between input and output.

\textit{Convolutional neural networks} (CNN) apply, besides other incredients, convolution kernels of different size in different layers in a sliding window approach to extract features. For a brief introduction to CNN, see \cite{Lecun2015}, e.g.
As an example, the prominent ResNet 50 network structure consists of 50 layers of convolutions or other layers, with \textit{skip connections} to avoid the problem of diminishing gradients.

\subsection{Transfer Learning}\label{sec:arch:TL}
Transfer learning (TL) deals with applying already gained knowledge for generalization to a different, but related domain~\cite{yosinski2014transferable}. Creating a separate, labeled dataset of sufficient size for a specific task of interest in the context of image classification is a time-consuming and resource-intensive process. Consequently, we find ourselves working with sets of training data that are significantly smaller than other renowned datasets, such as CIFAR and ImageNet~\cite{krizhevsky2009learning}. Moreover, the training process itself is time-consuming too and relies on dedicated hardware. Since modern CNNs take around 2-3 weeks to train on ImageNet in a professional environment, starting this process from scratch for every single model is hardly efficient. Therefore, general pretrained networks are typically used which are then tailored to specific inputs.

\section{Implementation}\label{sec:implementation}

\subsection{Automatic differentiation framework}
The \texttt{Newton-CG} optimization strategy is independent of the implementation, and of course, is suitable in any setting where second-order is beneficial \textbf{(1)} and storing Hessians is infeasible w.rr.t. memory consumption \textbf{(2)}. However, one needs a \textit{differentiation} framework. During the course of the work, a custom auto-encoder (and similar) implementation with optimized matrix operations\cite{chenhan2018distributed} became difficult, so with the abundance of Deep Learning around, we decided to move to a prominent framework, \textit{TensorFlow}. 
The TensorFlow programming model consists of two main steps: (1) Define computations in form of a ``stateful dataflow graph'' and (2) execute this graph. At the heart of model training in TensorFlow lies the Optimizer; we used \texttt{tf.python.keras.optimizer\_v2.Optimizer\_v2} subclassed like different optimizer algorithms (\texttt{Adam} or \texttt{SGD}).
The base class handles the two main steps of optimization: \texttt{compute\_gradients()}
and \texttt{apply\_gradients()}. When applying the gradients, for each variable that is optimized, the method \texttt{resource\_compute\_dense(grad, var)} is called with the
variable and its (earlier computed) gradient. In this method, the algorithm update step
for this variable is computed. It has to be overwritten by any subclassing optimizer.
We implemented two versions of our optimizer: one inheriting from the optimizer in tf.train and one inheriting from the Keras Optimizer\_v2. The constructor accepts the learning rate as well as the \texttt{Newton-CG} hyperparameters: regularization factor \(\tau\), the \texttt{CG}-convergence-tolerance and the maximum number
of CG iterations. Internally, the parameters are converted to tensors and stored as python object attributes.
The main logic happens in the above mentioned \texttt{resource\_compute\_dense(grad, var)}
method (see the implementation here\footnote{\url{{https://github.com/severin617/Newton-CG/blob/main/newton\_cg/newton\_cg.py\#L127}}}).

Table \ref{tab:5scen} lists the five ML scenarios and their implementation which have been used to generate the results below.
\begin{table}[]
\centering
\begin{tabular}{|l|l|}
\hline
scenario & \rule{.2em}{0em}description\\ \hline\hline
reg-lif   & \rule{.2em}{0em}one-layer life expectancy prediction\tablefootnote{ \url{https://valueml.com/predicting-the-life-expectancy-using-tensorflow/}}                \\
reg-bos   & \rule{.2em}{0em}two-layer boston housing price projection with keras \tablefootnote{ \url{https://www.kaggle.com/code/prasadperera/the-boston-housing-dataset}}           \\
vae-mnist & \rule{.2em}{0em}variational autoencoder from Keras  \tablefootnote{\url{https://keras.io/examples/generative/vae/}  }              \\
bnn-mnist & \rule{.2em}{0em}Bayesian neural network with tensorflow-probability\tablefootnote{\url{https://www.tensorflow.org/probability/}}~\cite{weigold2021} \\
resnet    & \rule{.2em}{0em}ResNet architecture from Keras  \tablefootnote{\url{https://www.tensorflow.org/api\_docs/python/tf/keras/applications/resnet50/ResNet50}}       \\
\hline        
\end{tabular}
\caption{\label{tab:5scen}Five ML scenarios with different neural network structures.}
\end{table}

%

\subsection{Data parallelism}
\label{subs:parallel}
In order to show the applicability of the proposed second-order optimizer for real-world large-scale networks, it was necessary to parallelize optimization computations to obtain suitable runtimes. We decided to use the comparably simple and prominent strategy of data parallelism. Data-parallel strategies t distribute data across different compute units, and each unit operates on the data in parallel. 
So in our setting, we compute different Newton-CG steps on $i$ different mini-batches in parallel, and the resulting update vectors are accumulated using an Allreduce. Note that this is different to e.g.~a $i$-times as big batch or $i$-times as many steps since this would use an updated weight when computing gradient information via backpropagations. In a smoothly defined function, this could converge to a similar minimum, however due to stochasticity this may not.  

 Horovod is a data-parallel distributed training framework (open source) for TensorFlow, Keras, PyTorch, and Apache MXNet, that scales a training script up to many GPUs using \texttt{MPI} \cite{sergeev2018horovod}. We apply Horovod for the data parallelisation of the second-order Newton-CG approach. In a second step the whole algorithms could be parallelized, this would then be model parallelism. 

The following table summarizes the data and model parallelism in the context of neural network optimization.


\vspace{.5em}
\begin{tabular}{p{5cm}|p{5cm}}
Data parallelism                                                 & Model parallelism                                                 \\ \hline \hline
Operations performed on different batches of data. & Parallel operations performed on same data (in identical batch).
\end{tabular}

\subsection{Software and hardware setup}

\begin{figure}[h!]
    \centering
    \includegraphics[width=0.6\textwidth]{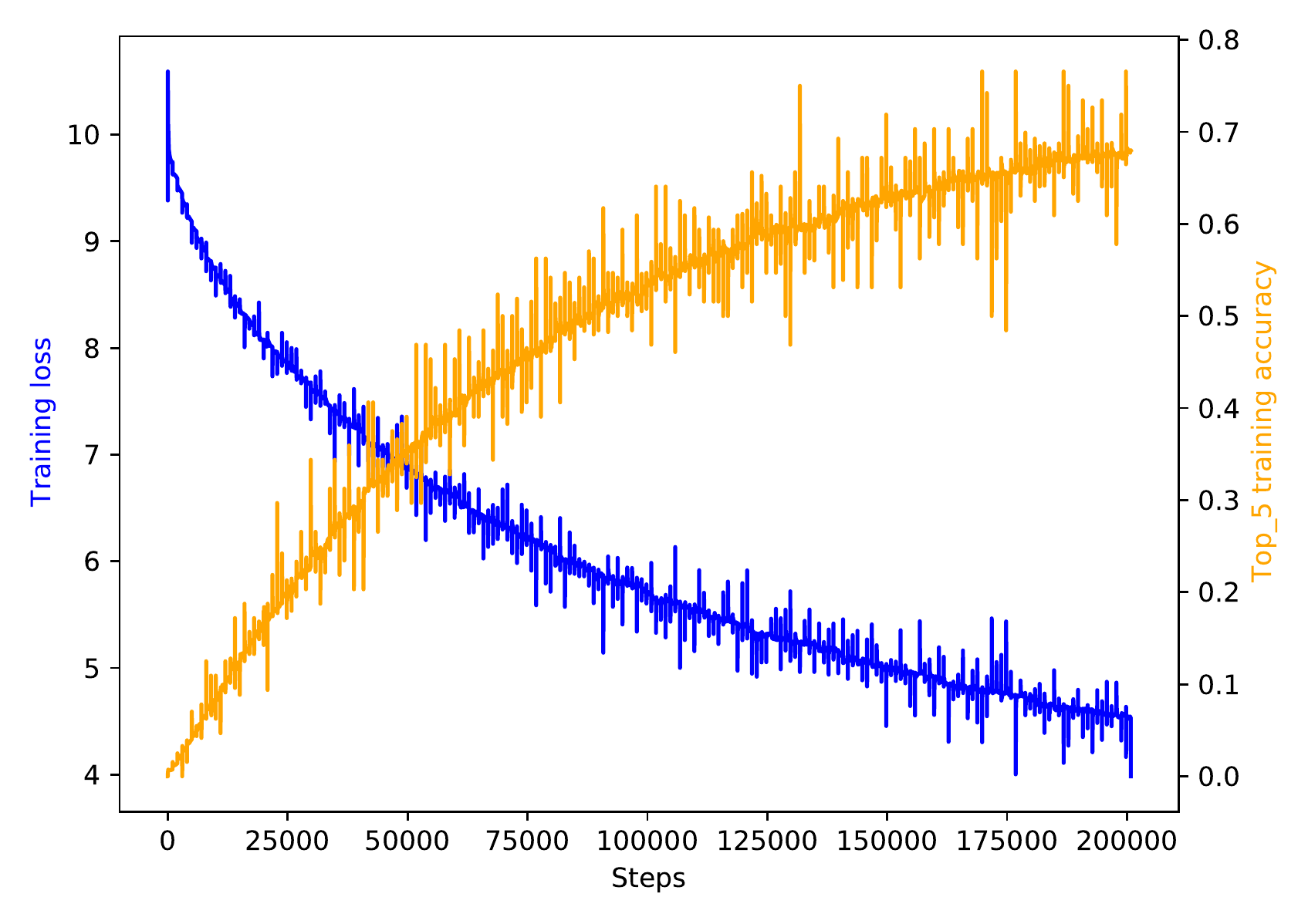}
    \caption{Training loss (blue) and accuracy (orange) of the SGD pretraining.}
    \label{fig:loss}
\end{figure}

Training with Keras and Horovod was used to show applicability and scalability of the proposed second order optimization.
The ResNets for computer vision were pretrained on a single-GPU machine with a NVIDIA Corporation GP102 [TITAN Xp] (rev a1). The training data from the Imagenet Large Scale Visual Recognition Challenge 2012 (ILSVRC2012) was used  with an SGD optimizer for 200 epochs \footnote{Following parameters were utilized in the pretraining: training/val-batch-size: 64, learning-rate: 0.001, momentum: 0.9, weight-decay: 0.00005. After each step, ten validation steps were used to calculate the top\_5 accuracy, resulting in a final loss of 4.5332 and a final top\_5 accuracy of 0.6800 after 2e5 steps.}.  
Figure \ref{fig:loss} shows the training loss and accuracy of these pretraining steps.
Test runs were performed on the \textit{Leibniz Rechenzentrum (LRZ)} AI System DGX-1 P100 Architecture with 8 NVIDIA Tesla P100 and 16 GB per GPU. 

\section{Results}\label{sec:results}

%

\subsection{Accuracy Results for Different Scenarios}

\begin{figure}[!ht]
  \centering
  \subfloat[][reg-lif]{\includegraphics[width=.48\textwidth]{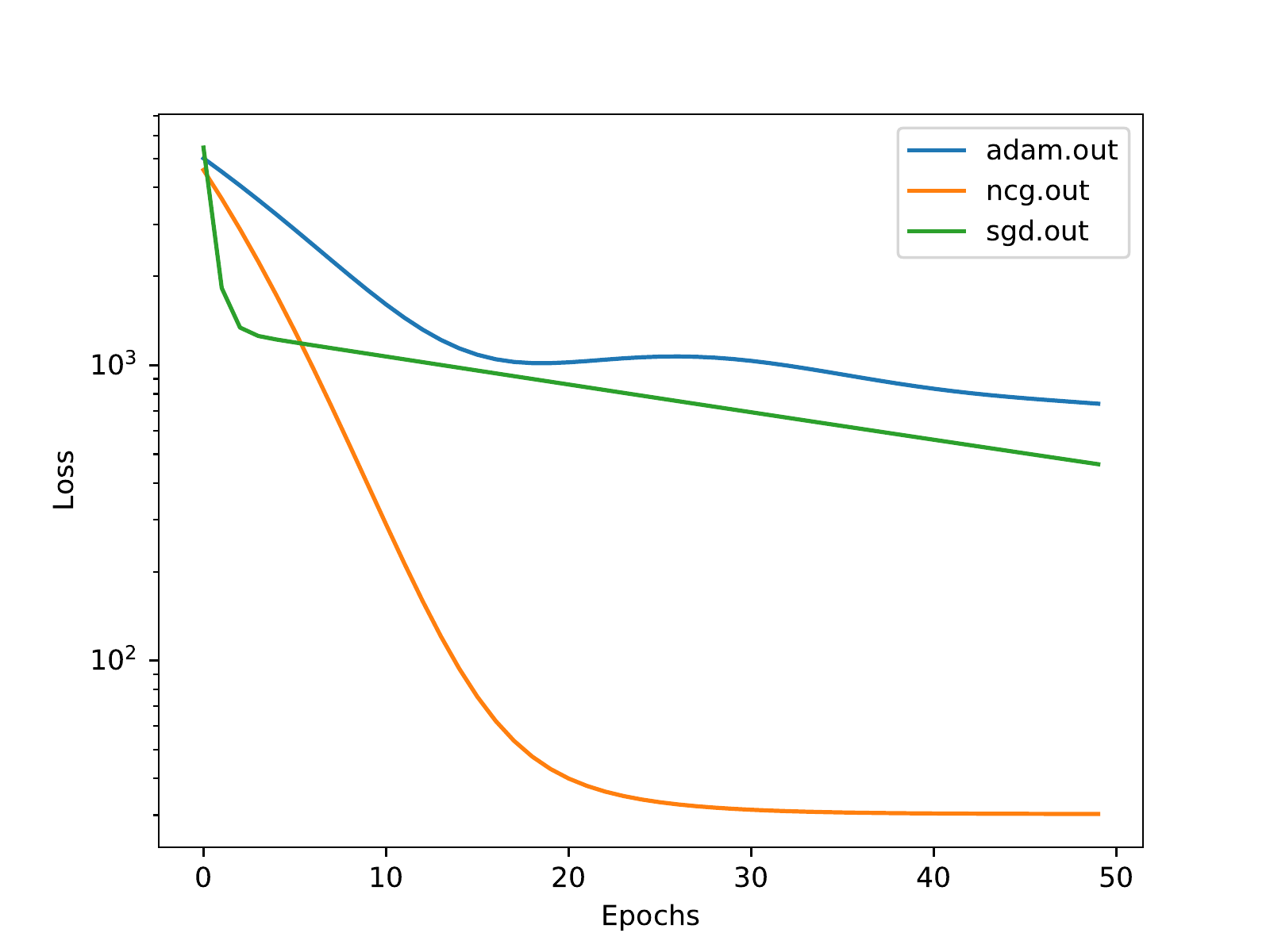}}\quad
  \subfloat[][reg-bos]{\includegraphics[width=.48\textwidth]{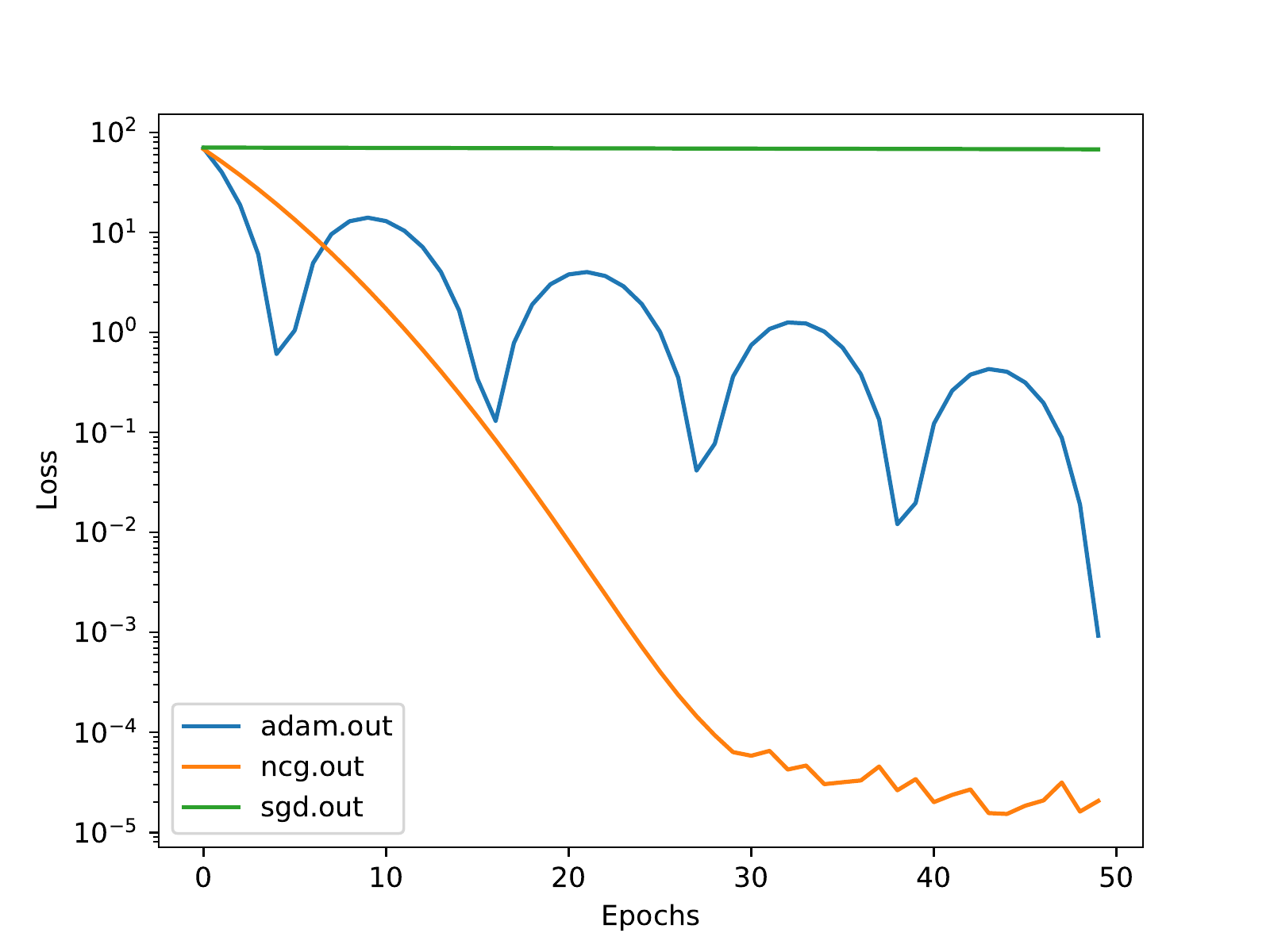}}\\
  \subfloat[][vae-mnist]{\includegraphics[width=.48\textwidth]{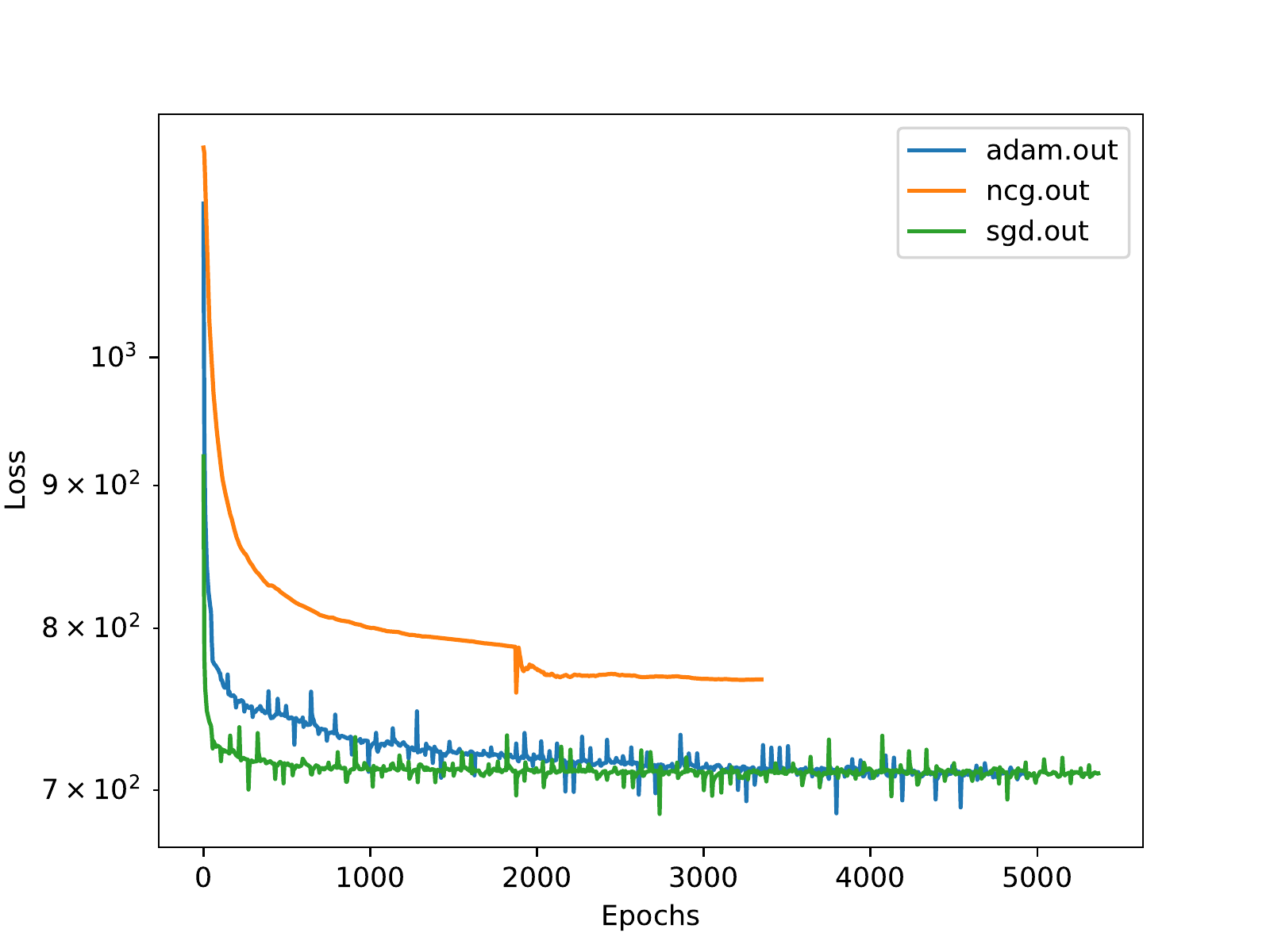}}\quad
  \subfloat[][bnn-mnist]{\includegraphics[width=.48\textwidth]{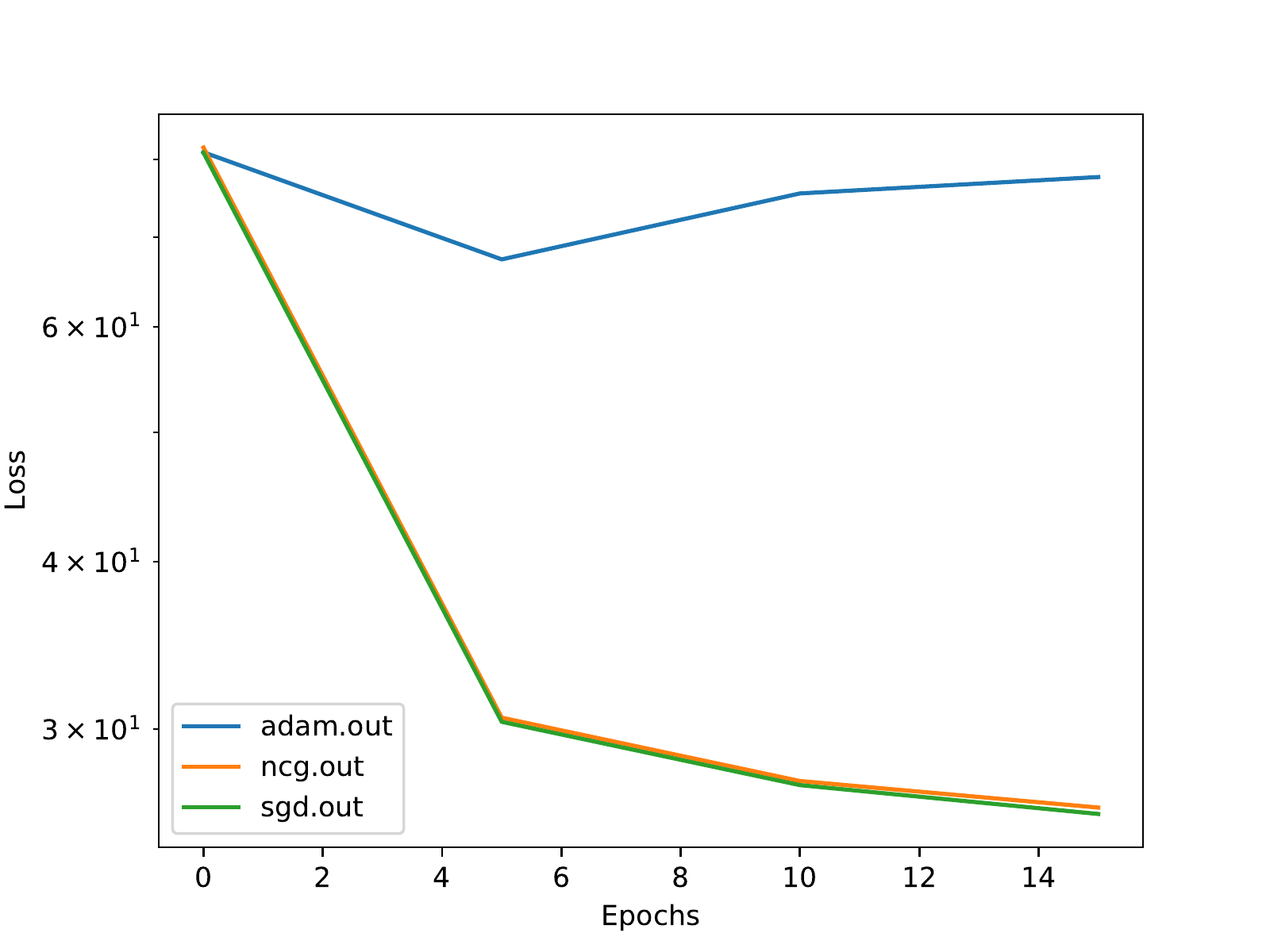}} \\
  \subfloat[][resnet]{\includegraphics[width=.48\textwidth]{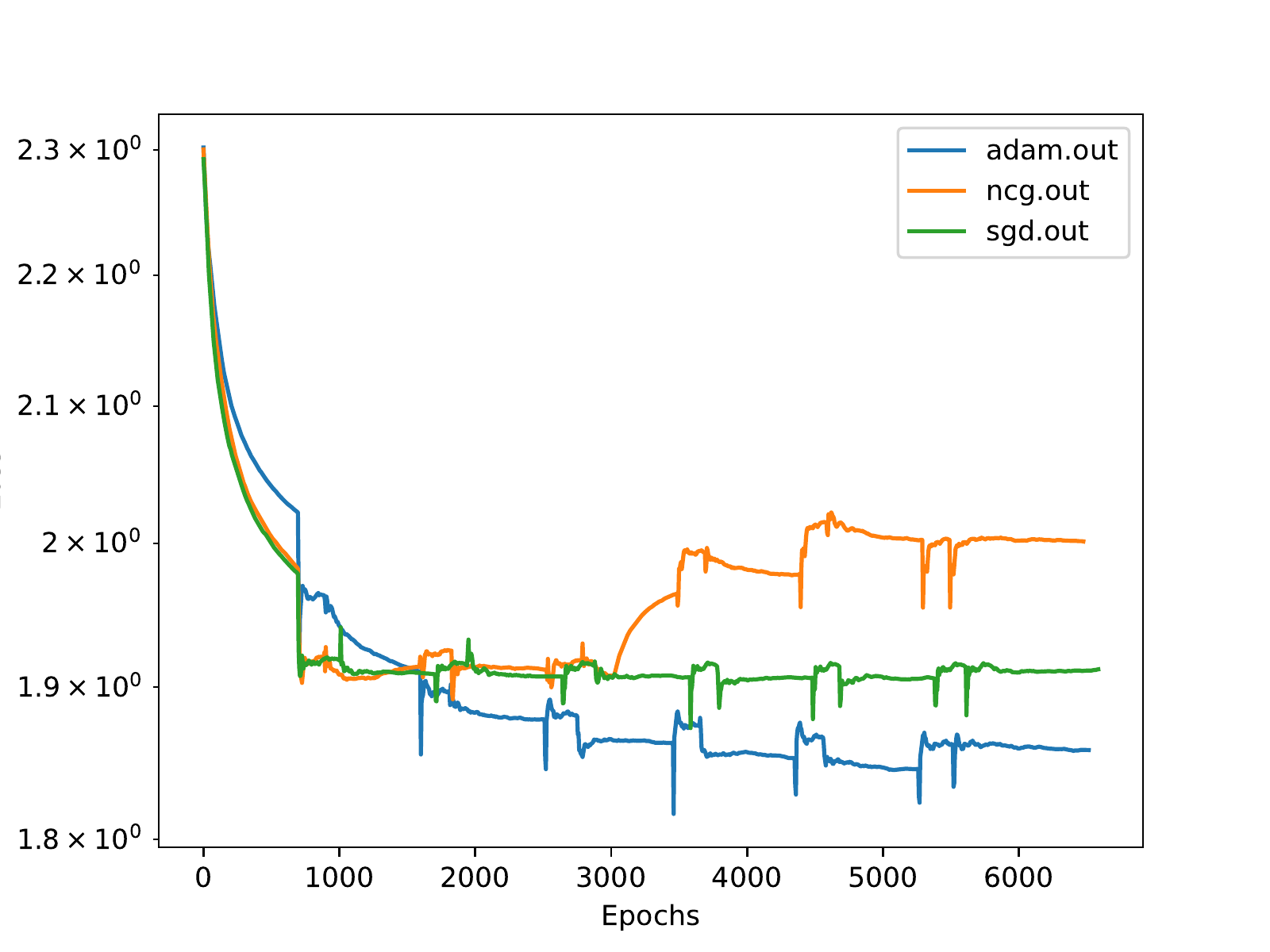}}
  \caption{Results for the training loss for the three compared methods: SGD in green, Adam in blue and Newton-CG in orange. The methods have been applied to the five different ML scenarios with corresponding different neural network structure: (a) regression case for life expectancy prediction, (b) regression for boston housing dataset, (c) Variational Auto Encoder with MNIST, (d) Bayesian Neural Network with MNIST, and (e) ResNet50 with ImageNet.}
  \label{fig:training}
\end{figure}

\begin{figure}[h!]
\centering
  \includegraphics[width=0.75\textwidth]{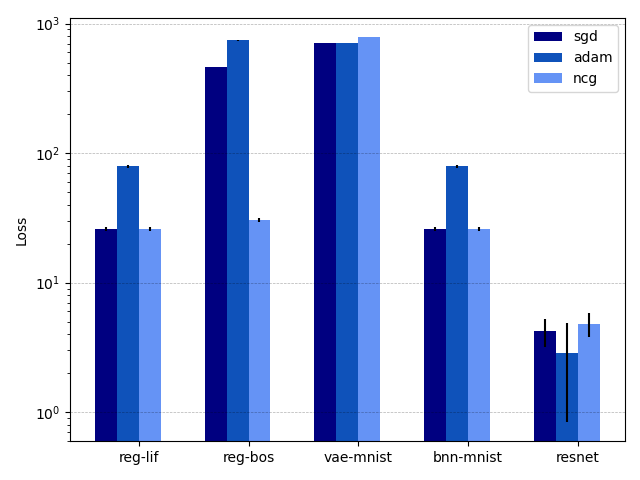}
  \caption{Final loss value of each optimizer for the five different neural network architectures and scenarios.}
  \label{fig:scenario}
\end{figure}

In this study, we applied the Newton-CG method as well as the two state-of-the-art methods SGD and Adam for the five different network architectures and specific scenarios described in Sec.~\ref{sec:arch} and Table \ref{tab:5scen} to evaluate the performance for each case and obtain insight into potential patterns. 
We show the detailed optimization behavior in Fig.~\ref{fig:training} while the final training loss optimum is summarized in Fig.~\ref{fig:scenario}. A similar comparison figure was used in \cite{schmidt2021descending} highlighting a similar insight that it is hard to predict the performance of different optimizers for considerably different scenarios. 

One can observe significant benefits of the 2nd-order Newton-CG in regression models, be it the life expectancy prediction or the boston housing data regression. We believe this is mostly due to the continuity in loss/optimization, whereas in the other scenarios this could jump, due to mini batches and classification. 

The variatonal autoencoder seems to work better with the conventional optimizers. Our hope was that due to the continuous behaviour we may see some benefits. However, this is also very hyper-parameter dependent, and the conventional methods have to be considerably tuned for that. In the Bayesian Neural Network we see benefits of Newton-CG especially against Adam. 

We observe hardly any benefits of 2nd-order optimization for the ResNet50 model. While at first we follow the near-optimal training curve, Newton-CG moves away from the minimum. One problem could be that we work with a fixed learning rate. This could be tuned with a \textit{learning-rate-scheduler}, which we currently work on.

\subsection{Parallel runs}
Exploiting parallelism allows for distributing work in case of failures (e.g. resilience), usage of modern compute architectures with accelerators,  and ultimately, lower \textit{time-to-solution}. 
All network architectures shown before can be run in parallel, in the data parallel approach explained in section \ref{subs:parallel}.

For the following measurements, we ran the ResNet50 model on the DGX-1 partition of the \textit{LRZ}, since it is our biggest network model and therefore, allows for the biggest parallelism gains (see Table \ref{tab:parallel}). \footnote{On the LRZ cluster, we had to reduce to 20\% training images for lower memory disk usage.}
 Note that the batch size is reduced with \textit{GPUs}, in order to account for a similar problem to be solved when increasing the amount of \textit{workers}.  However, it cannot be fully related to \textit{strong scaling}, since the algorithm changes as explained in Sec.~\ref{sec:implementation}. In a parallel setup, the loss is calculated for a smaller \textit{mini-batch} and then the \textit{update} is accumulated. This is different to looking at a bigger \textit{batch}, since the loss function is a different one.

\begin{table}[]
\caption{Newton-CG runtimes for 1 epoch with batch size 512, ResNet-50 with ImageNet}
\label{tab:parallel}
\centering
\begin{tabular}{c||c|c|c|c}
           & 1 GPU     & 2 GPUs     & 4  GPUs    & 8  GPUs    \\ \hline \hline
runtime    & 104s   & 60s     & 36s     & 23s     \\
parallel 
efficiency & 100\% & 86.6\% & 72.2\% & 56.5\%
\end{tabular}
\end{table}

\section{Conclusion and future work}\label{sec:conclusion}


In conclusion, we found benefits of second-order curvature information plugged into the optimization of the neural network weights especially for regression cases, but not much benefits in classification scenarios. In order to improve for classification, we experimented with a cyclical learning rate scheduler for ResNets for computer vision and Natural Language Processing, but more studies need to be investigated. The data-parallel approach seems to work well in performance numbers, since we reach about 56\% parallel efficiency for 8 GPUs.

For showcasing purposes, you may also try the  frontend android application TUM-lens\footnote{https://play.google.com/store/apps/details?id=com.maxjokel.lens }, where some models have been trained with Newton-CG.

\bibliographystyle{unsrt}

 \bibliography{bib}

\begin{thebibliography}{10}

\bibitem{Goodfellow-et-al-2016}
Ian Goodfellow, Yoshua Bengio, and Aaron Courville.
\newblock {\em Deep Learning}.
\newblock MIT Press, 2016.
\newblock \url{http://www.deeplearningbook.org}.

\bibitem{pearlmutter1994fast}
Barak~A Pearlmutter.
\newblock Fast exact multiplication by the hessian.
\newblock {\em Neural computation}, 6(1):147--160, 1994.

\bibitem{nocedal1999numerical}
Jorge Nocedal and Stephen~J Wright.
\newblock {\em Numerical optimization}.
\newblock Springer, 1999.

\bibitem{martens2010deep}
James Martens et~al.
\newblock Deep learning via hessian-free optimization.
\newblock In {\em ICML}, volume~27, pages 735--742, 2010.

\bibitem{martens2016second}
James Martens.
\newblock {\em Second-order optimization for neural networks}.
\newblock University of Toronto (Canada), 2016.

\bibitem{osawa2019large}
Kazuki Osawa, Yohei Tsuji, Yuichiro Ueno, Akira Naruse, Rio Yokota, and Satoshi
  Matsuoka.
\newblock Large-scale distributed second-order optimization using
  kronecker-factored approximate curvature for deep convolutional neural
  networks.
\newblock In {\em Proceedings of the IEEE/CVF Conference on Computer Vision and
  Pattern Recognition}, pages 12359--12367, 2019.

\bibitem{yao2020adahessian}
Zhewei Yao, Amir Gholami, Sheng Shen, Mustafa Mustafa, Kurt Keutzer, and
  Michael~W Mahoney.
\newblock Adahessian: An adaptive second order optimizer for machine learning.
\newblock {\em arXiv preprint arXiv:2006.00719}, 2020.

\bibitem{o2019inexact}
Thomas O'Leary-Roseberry, Nick Alger, and Omar Ghattas.
\newblock Inexact newton methods for stochastic nonconvex optimization with
  applications to neural network training.
\newblock {\em arXiv preprint arXiv:1905.06738}, 2019.

\bibitem{schmidt2021descending}
Robin~M Schmidt, Frank Schneider, and Philipp Hennig.
\newblock Descending through a crowded valley-benchmarking deep learning
  optimizers.
\newblock In {\em International Conference on Machine Learning}, pages
  9367--9376. PMLR, 2021.

\bibitem{chenhan2018distributed}
D~Yu Chenhan, Severin Reiz, and George Biros.
\newblock Distributed-memory hierarchical compression of dense spd matrices.
\newblock In {\em SC18: International Conference for High Performance
  Computing, Networking, Storage and Analysis}, pages 183--197. IEEE, 2018.

\bibitem{chen2021fast}
Chao Chen, Severin Reiz, Chenhan~D Yu, Hans-Joachim Bungartz, and George Biros.
\newblock Fast approximation of the gauss--newton hessian matrix for the
  multilayer perceptron.
\newblock {\em SIAM Journal on Matrix Analysis and Applications},
  42(1):165--184, 2021.

\bibitem{Lecun2015}
Yann Lecun, Yoshua Bengio, and Geoffrey Hinton.
\newblock {Deep learning}.
\newblock {\em Nature}, 521(7553):436--444, 2015.

\bibitem{shewchuk1994introduction}
Jonathan~Richard Shewchuk et~al.
\newblock An introduction to the conjugate gradient method without the
  agonizing pain, 1994.

\bibitem{suk}
Julian Suk.
\newblock Application of second-order optimisation for large-scale deep
  learning.
\newblock Masterarbeit, TUM, May 2020.

\bibitem{Kingma2019}
Diederik~P. Kingma and Max Welling.
\newblock {An Introduction to Variational Autoencoders}.
\newblock {\em Foundations and Trends{\textregistered} in Machine Learning},
  12(4):307--392, 2019.

\bibitem{bishop1995neural}
Christopher~M Bishop et~al.
\newblock {\em Neural networks for pattern recognition}.
\newblock Oxford university press, 1995.

\bibitem{blundell2015weight}
Charles Blundell, Julien Cornebise, Koray Kavukcuoglu, and Daan Wierstra.
\newblock Weight uncertainty in neural network.
\newblock In {\em International conference on machine learning}, pages
  1613--1622. PMLR, 2015.

\bibitem{yosinski2014transferable}
Jason Yosinski, Jeff Clune, Yoshua Bengio, and Hod Lipson.
\newblock How transferable are features in deep neural networks?
\newblock {\em Advances in neural information processing systems}, 27, 2014.

\bibitem{krizhevsky2009learning}
Alex Krizhevsky, Geoffrey Hinton, et~al.
\newblock Learning multiple layers of features from tiny images.
\newblock {\em Master's thesis, University of Tront}, 2009.

\bibitem{weigold2021}
Hanna Weigold.
\newblock Second-order optimization methods for bayesian neural networks.
\newblock Masterarbeit, Technischen Universität München, Jun 2021.

\bibitem{sergeev2018horovod}
Alexander Sergeev and Mike~Del Balso.
\newblock Horovod: fast and easy distributed deep learning in {TensorFlow}.
\newblock {\em arXiv preprint arXiv:1802.05799}, 2018.

\end{thebibliography}

 \end{document}


 \title{Training Arbitrary Neural Nets with a Newton Conjugate Gradient Method on Multiple GPUs  }
\titlerunning{Newton-CG for large ResNets} 
%
%
%
%
%
%
%
%
%
%
%
%
%
%
%

\appendix
\section{Appendix}

You may use Newton-CG freely, e.g. from github\footnote{https://github.com/severin617/Newton-CG} or get it from the Python Package Index (recommended with a new conda environment):
 \begin{minted}[fontsize=\footnotesize]{bash} 
 conda create -n tf1 python=3.7
 conda activate tf1
 pip install -r tensorflow==1.15 keras==2.3
 pip install -i https://test.pypi.org/simple/ newton-cg==0.0.3
  \end{minted}
You can use easily integrate it in any existing \texttt{tensorflow} or \texttt{keras} script, or with \texttt{tensorflow\_probability} for Bayesian Neural Networks:  
\begin{minted}[fontsize=\footnotesize]{python} 
import newton_cg as es
optimizer = es.EHNewtonOptimizer(
        learning_rate,
        tau=FLAGS.eso_tau,
        cg_tol=FLAGS.eso_cg_tol,
        max_iter=FLAGS.eso_max_iter)
 \end{minted}
 
 For parallelization, you can use the data-parallel approach of \texttt{horovod}, like this: 
 \begin{minted}[fontsize=\footnotesize]{python} 
 import newton_cg as es
 import horovod as hvd
 hvd.init()
 # Horovod: pin GPU to be used to process local rank (one GPU per process)
 config = tf.ConfigProto()
 config.gpu_options.allow_growth = True
 config.gpu_options.visible_device_list = str(hvd.local_rank())
 K.set_session(tf.Session(config=config))
 opt = EHNewtonOptimizer(initial_lr)
 es.opt = hvd.DistributedOptimizer(opt, compression=compression)
  \end{minted}
  Beware that alongside \texttt{horovod} setup with \texttt{MPI} and suitable \texttt{CUDA} drivers is tedious. You may get a docker image from NVIDIA directly\footnote{ngc.nvidia.com}, but we had to nevertheless make major adaptions here and there. We refer to the documentation and stackoverflow.
  
Especially the Computer Vision\footnote{https://gitlab.lrz.de/tum-i05/public/keras-second-order-optmizer} (\texttt{Imagenet}) and the NLP model\footnote{https://github.com/yunshu67/Bachelor-thesis---Extension-of-a-second-order-optimizer-to-NLP} (Portuguese-English Translation), requires large memory (and possibly small \texttt{batch-size}). Also, the program could need a long time to run due to \texttt{Imagenet} size (use pre-trained checkpoint file) or unrolling recurrent structures in NLP (\texttt{RNN}) despite using pre-trained word embeddings. To start off, use the regression, \texttt{MNIST} or Bayesian models.

%